# GTRL: An Entity Group-Aware Temporal Knowledge Graph Representation Learning Method

Xing Tang, Ling Chen

**Abstract**—Temporal Knowledge Graph (TKG) representation learning embeds entities and event types into a continuous low-dimensional vector space by integrating the temporal information, which is essential for downstream tasks, e.g., event prediction and question answering. Existing methods stack multiple graph convolution layers to model the influence of distant entities, leading to the over-smoothing problem. To alleviate the problem, recent studies infuse reinforcement learning to obtain paths that contribute to modeling the influence of distant entities. However, due to the limited number of hops, these studies fail to capture the correlation between entities that are far apart and even unreachable. To this end, we propose GTRL, an entity Group-aware Temporal knowledge graph Representation Learning method. GTRL is the first work that incorporates the entity group modeling to capture the correlation between entities by stacking only a finite number of layers. Specifically, the entity group mapper is proposed to generate entity groups from entities in a learning way. Based on entity groups, the implicit correlation encoder is introduced to capture implicit correlations between any pairwise entity groups. In addition, the hierarchical GCNs are exploited to accomplish the message aggregation and representation updating on the entity group graph and the entity graph. Finally, GRUs are employed to capture the temporal dependency in TKGs. Extensive experiments on six real-world datasets demonstrate that GTRL achieves the state-of-the-art performances on the event prediction task, outperforming the best baseline by an average of 7.35%, 6.09%, 8.31%, and 11.21% in MRR, Hits@1, Hits@3, and Hits@10, respectively.

**Index Terms**—Temporal knowledge graph, representation learning, entity group modeling, graph convolution network

✦ ⸺⸺⸺⸺⸺⸺⸺⸺⸺⸺

## 1 INTRODUCTION

TEMPORAL Knowledge Graphs (TKGs) [23], e.g., the Global Database of Events, Language and Tone (GDELT) [15] and the Integrated Conflict Early Warning System (ICEWS) [2], include a large amount of temporally annotated knowledge, i.e., events. Events in TKGs are denoted as (head entity, event type, tail entity, timestamp), abbreviated as quadruplet $(s, r, o, t)$. For example, for event (New Zealand, Express intent to cooperate, Greece, 2018/1/7), "New Zealand" and "Greece" refer to head entity and tail entity, respectively, "Express intent to cooperate" refers to event type, and "2018/1/7" refers to timestamp. TKG representation learning [28] aims to embed entities and event types into a continuous low-dimensional vector space by incorporating the temporal information, which is a common way to represent entities and event types and is of significance for downstream tasks, e.g., event prediction [14] and question answering [6]. Compared with static Knowledge Graph (KG) representation learning [11], TKG representation learning is non-trivial and faces the challenge of how to effectively utilize the temporal information.

Existing TKG representation learning methods can be classified into two categories [10], i.e., interpolation-based methods and extrapolation-based methods. Interpolation-based methods focus on completing missing events in the history and extending static KG representation learning methods to TKGs, which shuffle events randomly and straightly model timestamps as hyperplanes [5], representations [14], or the order of tensor decomposition [13]. However, these methods cannot model the temporal dependency in TKGs, in which timestamps are simply encoded in a low-dimensional vector space.

Extrapolation-based methods aim at predicting events in the future, which infuse deep neural networks to model the temporal dependency. Some studies introduce Recurrent Neural Networks (RNNs) and their variants, e.g., Long Short-Term Memory (LSTM) and Gated Recurrent Units (GRUs), as well as self attention, to model the temporal dependency, and incorporate Graph Convolution Networks (GCNs) to model the graph structure information [7], [10], [26], [32]. In these studies, entity graphs are constructed based on historical events and entity representations are learned by aggregating the message from neighbor entities. In fact, at a specific time step, some entities are far apart and even unreachable on the entity graph, but have strong correlations. In Fig. 1, we provide an example of entities that are far apart. Although "Dominican Republic" and "China" are far apart on the entity graph constructed from ICEWS on January 12, 2018, "China" is important in portraying "Dominican Republic" due to the Belt and Road Initiative. In addition, there is probably no reachable path between some entities, but

⸺⸺⸺⸺⸺⸺⸺⸺

- *This work was supported in part by the National Key Research and Development Program of China under Grant 2018YFB0505000 and in part by the project of the Donghai Laboratory under Grant DH-2022ZY0013. (Corresponding author: Ling Chen).*
- *Xing Tang and Ling Chen are with the State Key Laboratory of Blockchain and Data Security, Zhejiang University, Hangzhou 310027, China, and also with the College of Computer Science and Technology, Zhejiang University, Hangzhou 310027, China (e-mail: tangxing@cs.zju.edu.cn; lingchen@cs.zju.edu.cn).*

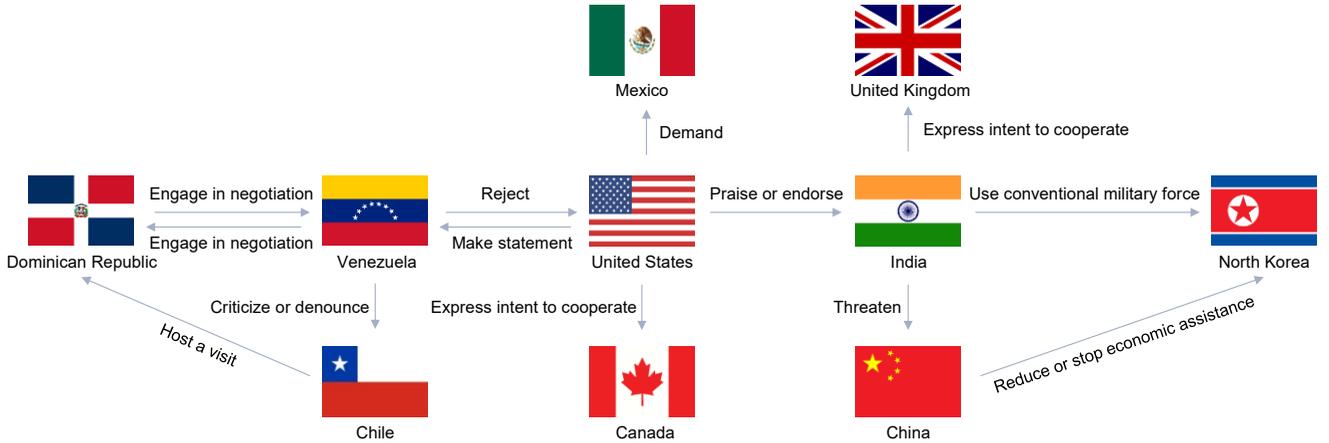

Fig. 1. Fragment of the entity graph constructed from ICEWS on January 12, 2018. (Ten countries are selected, i.e., Mexico, United Kingdom, Dominican Republic, Venezuela, United States, India, North Korea, Chile, Canada, and China. If there are many events from one country to another, only one is retained for brevity.)

they are still highly correlated. To model the influence of distant entities, existing methods often stack multiple graph convolution layers [7], [10]. However, stacking multiple graph convolution layers leads to the over-smoothing problem [27], i.e., the representations of neighbor entities will become similar.

In order to alleviate the problem, another line of work models the influence of distant entities based on paths, which are obtained by reinforcement learning [8], [9]. In spite of the promising results of introducing paths, there still leaves much to be desired. Since the number of paths increases exponentially with the number of hops, these methods usually model 2-hop or 3-hop paths, neglecting rich information from longer paths, which fail to capture the correlation between entities that are far apart and even unreachable.

To address the above-mentioned drawback, we propose GTRL, an entity Group-aware Temporal knowledge graph Representation Learning method. To the best of our knowledge, GTRL is the first work that incorporates the entity group modeling into TKG representation learning, which can model the influence of distant entities by stacking only a finite number of layers, thus avoiding the over-smoothing problem caused by stacking multiple layers in existing methods. The main contributions of this paper are outlined as follows:

1) We propose the entity group mapper to generate entity groups from entities in a learning way. In addition, the implicit correlation encoder is introduced to capture implicit correlations between any pairwise entity groups.

2) We introduce the hierarchical GCNs to implement the information interaction on the entity group graph and the entity graph, which can model the influence of both adjacent and distant entities.

3) We conduct extensive experiments on six real-world datasets. Experimental results demonstrate that GTRL achieves the state-of-the-art (SOTA) performances on the event prediction task, outperforming the best baseline by an average of 7.35%, 6.09%, 8.31%, and 11.21% in MRR, Hits@1, Hits@3, and Hits@10, respectively.

The remainder of this paper is structured as follows: We provide a review of existing methods associated with this work in Section 2. We provide the definitions of the associated terms and give the formulation of the task in Section 3. We present the specifics of GTRL in Section 4. The experimental results and analyses are given in Section 5. Finally, we conclude the paper and present the future work in Section 6.

## 2 RELATED WORK

This section provides an overview of the related work, including static KG representation learning methods, TKG representation learning methods, and temporal graph representation learning methods.

### 2.1 Static KG Representation Learning Methods

Static KG representation learning methods can be generally divided into three classes, i.e., translation methods, semantic matching methods, and GCN-based methods. TransE [1] is a typical translation method, which assumes that the tail entity representation should be close to the sum of the head entity representation and the relation representation. TransE has an impressive performance in handling 1-to-1 relations, but has troubles in handling 1-to-N, N-to-1, and N-to-N relations. To address the problem, plenty of variants based on TransE are proposed [12], [17], [22], [25], [33], [34]. RESCAL [18] is a typical semantic matching method, which models semantic correlations between entities and relations by a bilinear model. DistMult [29] simplifies RESCAL by employing diagonal matrices to represent relations, which reduces the number of parameters. As a common GCN-based method, R-GCN [20] assigns different weights for relations and obtains entity representations by utilizing GCNs to model the influence of neighbor entities. SACN [21] distinguishes the influence of different neighbor entities based on the weighted GCNs. CompGCN [24] learns the representa-tions of both entities and relations by introducing entity-relation composition operations. In contrast to translation methods and semantic matching methods that just use triplets, i.e., (head entity, relation, tail entity), GCN-based methods exploit the graph structure information, which contributes to learning more effective representations.

**2.2 TKG Representation Learning Methods**

TKG representation learning, integrating the temporal information, aims to embed entities and event types into a continuous low-dimensional vector space, which has received wide attention. Existing TKG representation learning methods can be classified into two categories [10], i.e., interpolation-based methods and extrapolation-based methods. Interpolation-based methods aim to complete missing events in the history. The interpolation task focuses on filling in the blanks within a known time range. Extrapolation-based methods aim to predict events in the future, which infuse deep neural networks to model the temporal dependency. The extrapolation task focuses on predicting the future based on known historical data. While extrapolation-based methods are mainly designed for predicting future events, they might also be applicable to the interpolation task in some scenarios. However, their performance might not be as effective as interpolation-based methods specifically designed for the interpolation task. Generally, interpolation-based methods are not applied for the extrapolation task, as they fail to consider the temporal dependency and potential evolution patterns in the future.

Interpolation-based methods extend static KG representation learning methods to TKGs, which shuffle events randomly and directly encode corresponding timestamps in a simple way. HyTE [5], TTransE [14], and TNTComplEx [13] model timestamps in the way of the hyperplane projection, representation encoding, and tensor decomposition, respectively. However, these methods fail to model the temporal dependency in TKGs.

Extrapolation-based methods predict future events based on historical data, which employ deep neural networks to model the temporal dependency. Some studies introduce RNNs and their variants, e.g., LSTM and GRUs, as well as self attention, to model the temporal dependency, and incorporate GCNs to model the graph structure information. RE-NET [10] exploits GCNs to obtain entity representations by modeling the influence of neighbor entities, employs RNNs to model the temporal dependency among events, and obtains the joint probability distribution of all events in an autoregressive way. Glean [7] employs CompGCN to model the influence of neighbor entities and event types, and adopts GRUs to model the temporal dependency among representations. TeMP [26] uses R-GCN to model the influence of neighbor entities and exploits a frequency-based gating GRU to model the temporal dependency among inactive events. DACHA [30] introduces a dual graph convolution network to obtain entity representations, which models the information interaction on the primal graph and the edge graph, and utilizes a self-attentive encoder to model the temporal dependency among event types. RE-GCN [32] uses R-GCN to aggregate the message from neighbor entities and employs an auto-regressive GRU to model the temporal dependency among events. However, these methods often stack multiple graph convolution layers to model the influence of distant entities, which leads to the over-smoothing problem [27], i.e., the representations of neighbor entities tend to become similar.

To alleviate the over-smoothing problem, recent studies infuse reinforcement learning to obtain paths that contribute to modeling the influence of distant entities [8], [9]. TITer [8] incorporates temporal agent-based reinforcement learning to search paths and obtains entity representations by inductive mean. CluSTeR [9] introduces beam search policy-based reinforcement learning to search paths and incorporates GCNs and GRUs to learn entity representations. Although these methods have achieved promising results by introducing paths, there still leaves much to be desired. Since the number of paths increases exponentially with the number of hops, these methods usually model paths with 2 or 3 hops, neglecting rich information from longer paths, which fail to capture the correlation between entities that are far apart and even unreachable.

To tackle the aforementioned drawback, we propose GTRL, an entity group-aware temporal knowledge graph representation learning method. Empowered by the entity group modeling, GTRL can capture the correlation between entities that are far apart and even unreachable. The dedicated design allows GTRL to model the influence of distant entities by stacking only a finite number of layers, thus avoiding the over-smoothing problem caused by stacking multiple layers in existing methods.

**2.3 Temporal Graph Representation Learning Methods**

Temporal graph representation learning methods aim to capture node evolutionary patterns and accurately predict future links, which have been widely applied in real-world scenarios, e.g., recommendation and social networks. Existing methods can be mainly divided into two categories: Discrete-time temporal graph embedding methods and continuous-time temporal graph embedding methods. EvolveGCN [40] is a typical discrete-time temporal graph embedding method, which introduces RNNs to model the parameter dynamics of the GNNs used in adjacent network snapshots. However, manually selecting the time window hinders the modeling of more intricate temporal patterns.

To handle this problem, continuous-time temporal graph embedding methods have been proposed. JODIE [41] leverages RNNs to chronologically update the representations of interconnected nodes based on the sequence of their interactions. TigeCMN [42] uses memory networks to update node representations sequentially based on the attention mechanism. To integrate the overall topology structure, TGAT [43] exploits the self-attention mechanism to obtain time-augmented node representations, and employs GNNs to aggregate multi-hop neighbors. TGN [44] captures the temporal information using RNNs, and applies graph attention convolution to capture the spatial and temporal information jointly. APAN [45] employs a mailbox mechanism to store adjacent states for each node, enabling local temporal graph aggregation during online inference, which enhances inference speed through a space-time tradeoff. However, these methods predominantly focus on heterogeneous graphs and tend to falter when dealing with the complex entities and relations in heterogeneous graph scenarios like TKGs.

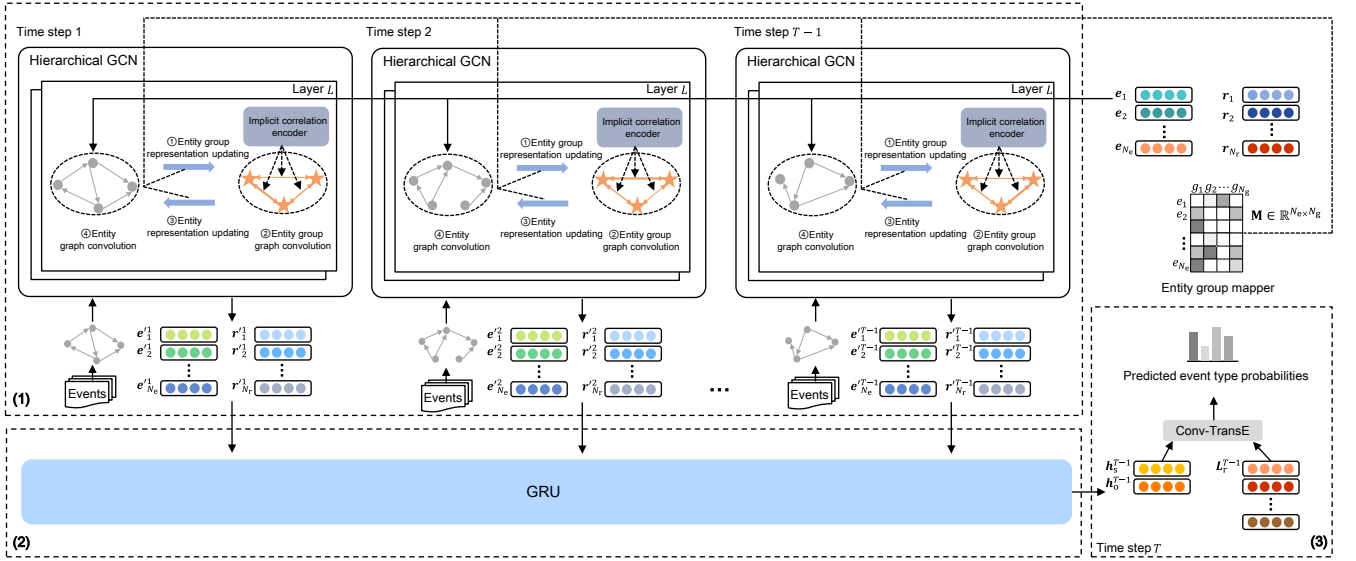

Fig. 2. Framework of GTRL: (1) Interaction modeling; (2) Temporal dependency modeling; (3) Event prediction.

## 3 DEFINITIONS AND PRELIMINARIES

In this section, we provide the definitions of the associated terms used in GTRL, and formulate the event prediction task in TKGs.

**Definition 1**. **TKG**. A TKG can be formalized as a set of events, denoted as $\mathcal{G} = \{(s,r,o,\tau) | s, o \in \mathcal{E}, r \in \mathcal{R}, \tau \in \mathcal{T}\}$, where $\mathcal{E}$, $\mathcal{R}$, and $\mathcal{T}$ denote the sets of entities, event types, and timestamps, respectively. The set of events at time step $t$ is denoted as $\mathcal{G}^t = \{(s,r,o,\tau)\}^t$.

**Definition 2**. **Entity Graph**. The entity graph is regarded as a multi-relational and directed graph, denoted as $G_{\text{entity}} = (V_{\text{entity}}, E_{\text{entity}})$, where $V_{\text{entity}}$ and $E_{\text{entity}}$ denote the sets of nodes and edges on the entity graph, respectively. The nodes and edges represent entities and events between them, denoted as $e$ ($s$ or $o$) and $r$, respectively. Each edge has a direction pointing from head entity to tail entity.

**Definition 3**. **Entity Group Graph**. The entity group graph is regarded as a fully connected graph, denoted as $G_{\text{entity group}} = (V_{\text{entity group}}, E_{\text{entity group}})$, where $V_{\text{entity group}}$ and $E_{\text{entity group}}$ denote the sets of nodes and edges on the entity group graph, respectively. The nodes and edges represent entity groups and implicit correlations between them, denoted as $g$ and $c$, respectively.

**Problem 1**. **Event Prediction Task**. The event prediction task in TKGs aims to predict the probability of an event type between an entity pair at time step $T$, based on historical events at previous $T-1$ time steps, formulated as $P(r|s, o, \mathcal{G}^{1:T-1})$, where $\mathcal{G}^{1:T-1}$ is the set of historical events at previous $T-1$ time steps, $s$ and $o$ are the given entity pair, and $r$ is the event type at time step $T$.

## 4 METHODOLOGY

In this section, we firstly give the framework of GTRL and then describe individual modules in detail.

### 4.1 Overview

The framework of GTRL is shown in Fig. 2, which consists of three modules: (1) Interaction modeling; (2) Temporal dependency modeling; (3) Event prediction. For the interaction modeling module, the entity graph is built based on historical events for each time step in the historical window $[1:T-1]$. The entity group mapper generates entity groups from entities. The implicit correlation encoder captures implicit correlations between anypair-wise entity groups. In addition, the hierarchical GCNs accomplish the message aggregation and representation updating on the entity group graph and the entity graph, which can model the influence of both adjacent and distant entities. For the temporal dependency modeling module, GRUs capture the temporal dependency among updated representations at previous $T-1$ time steps. For the event prediction module, final entity representations and the event type representation matrix are input to the Conv-TransE, which predicts the probability of each event type at time step $T$.

GTRL seamlessly integrates the entity group modeling into TKGs. This integration is achieved by combining hierarchical GCNs with GRUs to capture both graph structure information and temporal information.

## 4.2 Interaction Modeling

### 4.2.1 Entity Group Representation Updating

To utilize the crucial information from historical events to depict entities, the entity graph is constructed based on historical events for each time step in the historical window $[1:T-1]$. The edge construction rule of the entity graph is formulated as:

$$r_{i,j} = \begin{cases} 1, \text{there is an event from entity } e_i \text{ to entity } e_j \\ 0, \text{otherwise} \end{cases} \quad (1)$$

Each entity $e_i$ and event type $r_i$ are given with initialized representations $\boldsymbol{e}_i$ and $\boldsymbol{r}_i$, respectively.

Distant entities may have strong correlations, but stacking multiple graph convolution layers to model the influence of distant entities would cause the over-smoothing problem. Thus, the entity group mapper is designed to capture the mapping between entities and entity groups in a learning way. Instead of assigning each entity into a specific entity group, the entity group mapper makes each entity belong to multiple entity groups with different probabilities.

Specifically, the entity group mapper utilizes mapping matrix $\mathbf{M} \in \mathbb{R}^{N_e \times N_g}$ to capture the mapping between entities and entity groups, where $N_e$ and $N_g$ are the numbers of entities and entity groups, respectively. To reduce the impact of noise and make the model robust, we introduce a threshold strategy to make $\mathbf{M}$ sparse, which utilizes the sparsemax function to adaptively preserve the values above the threshold that is calculated based on $\mathbf{M}$ [37], and the other values are truncated to 0. $\mathbf{M}$ is shared over all time steps. Its parameters are randomly initialized and would be optimized during the training process. The entity group representation is formulated as:

$$\boldsymbol{g}_j = \sum_{e_i \in V_{\text{entity}}} \mathbf{M}_{i,j} \boldsymbol{e}_i \quad (2)$$

where $\mathbf{M}_{i,j}$ is the probability of assigning entity $e_i$ to entity group $g_j$, $i \in \{1,2,\ldots,N_e\}$, $j \in \{1,2,\ldots,N_g\}$, and $\sum_{j=1}^{N_g} \mathbf{M}_{i,j} = 1$. $V_{\text{entity}}$ is the set of entities.

### 4.2.2 Entity Group Graph Convolution

Capturing implicit correlations between entity groups is of significance and helps to make an accurate prediction. Thus, we introduce the implicit correlation encoder to capture implicit correlations between entity groups in an end-to-end way. To avoid missing any relevant information between entity groups, the entity group graph is modeled as the fully connected graph. Given that the number of entity groups is much smaller than that of entities, we assume that there is an implicit correlation between any pairwise entity groups. The representation of the implicit correlation between entity groups is obtained based on entity group representations, which is formulated as:

$$\boldsymbol{c}_{i,j} = \text{ReLU}(\varphi(\boldsymbol{g}_i, \boldsymbol{g}_j)) \quad (3)$$

where $\boldsymbol{g}_i$ and $\boldsymbol{g}_j$ are entity group representations, calculated by Equation (2), and $i,j \in \{1,2,\ldots,N_g\}$. $\varphi$ is the transformation function, implemented by the multi-layer perceptron (MLP).

To distinguish different implicit correlations, the representation of the implicit correlation is used to obtain the intensity of the implicit correlation, which is formulated as:

$$q_{i,j} = \sigma(\text{Conv}(\boldsymbol{c}_{i,j})) \quad (4)$$

where $\text{Conv}(\cdot)$ represents the 1D convolution layer. $\sigma$ is the sigmoid function, which keeps the value of the intensity of the implicit correlation ranging from 0 to 1.

In the entity group graph convolution, the stronger the intensity of the implicit correlation between an entity group pair is, the more critical the implicit correlation is in the message aggregation process and vice versa. Thus, GTRL aggregates messages from connected edges based on the intensity of the implicit correlation. The aggregating operation is formulated as:

$$\boldsymbol{a}_{g_i} = \sum_{i \neq j} q_{i,j} \boldsymbol{c}_{i,j} \quad (5)$$

where $q_{i,j}$ and $\boldsymbol{c}_{i,j}$ are the intensity and the representation of the implicit correlation calculated by Equation (4) and Equation (3), respectively. Then, the entity group representation is updated, which is formulated as:

$$\boldsymbol{g}'_i = \varphi(\boldsymbol{a}_{g_i}, \boldsymbol{g}_i) \quad (6)$$

where $\varphi$ is the transformation function, implemented by the MLP.

### 4.2.3 Entity Representation Updating

To implement the information interaction from the entity group graph to the entity graph, the mapping matrix (detailed in Section 4.2.1) is adopted to update the entity representation based on corresponding entity group representations, which is formulated as:

$$\boldsymbol{e}_i = \sum_{j=1}^{N_g} \mathbf{M}_{i,j} \boldsymbol{g}'_j \quad (7)$$

where $N_g$ is the number of entity groups. $\boldsymbol{g}'_j$ is the updated entity group representation, calculated by Equation (6).

### 4.2.4 Entity Graph Convolution

In the entity graph convolution, GTRL aggregates messages from neighbor entities by the aggregating operation and the updating operation. The aggregating operation is formulated as:

$$a_{e_i} = \varrho(\{e_i, e_j, r_{i,j}\}_{e_j \in \mathbb{N}_{e_i}}) \quad (8)$$

where $\varrho$ is the aggregation function, implemented by the CompGCN [24], an advanced static KG representation learning method for learning the representations of entities and relations. $\mathbb{N}_{e_i}$ is the neighbor set of entity $e_i$ on the entity graph. Then, the entity representation is updated, which is formulated as:

$$e'_i = \varphi(a_{e_i}, e_i) \quad (9)$$

where $\varphi$ is the transformation function, implemented by the MLP.

Subsequently, the event type representation is also updated, which is formulated as:

$$r'_i = W_r r_i \quad (10)$$

where $W_r$ is the learnable parameter, which projects the event type representation to the same vector space as the entity representation.

In this way, the hierarchical GCNs output updated entity representations and updated event type representations at previous $T-1$ time steps, which can be formulated as $e'^{1:T-1}_i = \{e'^1_i, e'^2_i, \ldots, e'^t_i, \ldots, e'^{T-1}_i\}$ and $r'^{1:T-1}_i = \{r'^1_i, r'^2_i, \ldots, r'^t_i, \ldots, r'^{T-1}_i\}$, respectively. The updated representations integrate the neighbor information on both the entity graph and the entity group graph, which can model the influence of both adjacent and distant entities. Note that if no event occurs for an entity or an event type at a certain time step, the entity representation or the event type representation at the previous time step will be retained.

### 4.3 Temporal Dependency Modeling

Since GRUs have shown advantages in modeling the temporal dependency, following [16], we leverage GRUs to model the temporal dependency among updated representations. In fact, if an event has occurred for a long time, its influence is limited. To capture the importance variation of events in the history, the decay rate [4] is introduced. Note that for simplicity, the subscript $i$ of variables is omitted in the following without causing ambiguity. The decay rate is formulated as:

$$\gamma^t = \sigma\{-\max(0, W_\gamma |t - t'| + b_\gamma)\} \quad (11)$$

where $W_\gamma$ and $b_\gamma$ are learnable parameters. $t'$ is the time step of the last event before time step $t$ for the entity. $\sigma$ is the sigmoid function, which keeps the value of the decay rate monotonically decreasing in a reasonable range between 0 and 1.

Considering the decay of the previous hidden state, the new hidden state $\hat{h}^{t-1}$ is formulated as:

$$\hat{h}^{t-1} = \gamma^t h^{t-1} \quad (12)$$

The updating process of GRUs is formulated as:

$$\begin{aligned} r^t &= \sigma(W_r[X^t, \hat{h}^{t-1}] + b_r) \\ z^t &= \sigma(W_z[X^t, \hat{h}^{t-1}] + b_z) \\ \tilde{h}^t &= \tanh(W_{\tilde{h}}[X^t, r^t \odot \hat{h}^{t-1}] + b_{\tilde{h}}) \\ h^t &= z^t \odot \hat{h}^{t-1} + (1 - z^t) \odot \tilde{h}^t \end{aligned} \quad (13)$$

where $W_r$, $W_z$, $W_{\tilde{h}}$, $b_r$, $b_z$, and $b_{\tilde{h}}$ are learnable parameters. $X^t$ denotes the input at time step $t$, which is the updated entity representation or the updated event type representation output by the hierarchical GCNs. $h^t$ denotes the output at time step $t$. $r^t$, $z^t$, and $\tilde{h}^t$ denote the reset gate, update gate, and new gate at time step $t$, respectively. $[\cdot]$ denotes the concatenation operation. $\sigma$ is the sigmoid function, and $\odot$ is the Hadamard product.

In this way, the output of GRUs, i.e., the final entity representation and the event type representation at time step $T-1$, formulated as $h_e^{T-1}$ ($h_s^{T-1}$ or $h_o^{T-1}$) and $h_r^{T-1}$, respectively, integrate the temporal information and neighbor information. It is worth noting that GRUs can be replaced with any advanced sequence model, e.g., temporal self attention [19] and temporal GRUs [26], which are augmented with customized mechanism to model different types of temporal evolution in entities, for modeling the temporal dependency among updated representations.

### 4.4 Event Prediction

Following [32], we employ the Conv-TransE [21] to predict the probability of each event type between an entity pair. Specifically, final entity representations are stacked and input to the 1D convolutional layer to obtain feature maps via multiple convolution kernels of the same size. Then, feature maps are vectorized and input to the fully connected layer to obtain the representation, which has the same dimension as the original representation. Finally, the representation is multiplied by the final event type representation matrix to obtain the probability of each event type via the sigmoid function. The probability of each event type between an entity pair is calculated as:

$$P(\hat{y}|s, o, \mathcal{G}^{1:T-1}) = \sigma(\mathbf{L}_r^{T-1} \text{ConvFC}(h_s^{T-1}, h_o^{T-1})) \quad (14)$$

where $\hat{y} \in \mathbb{R}^{|\mathcal{R}|}$ is the vector of event types, and $|\mathcal{R}|$ is the number of event types. $\sigma$ is the sigmoid function. $h_s^{T-1}$ and $h_o^{T-1}$ are final entity representations at time step $T-1$. $\mathbf{L}_r^{T-1}$ is the final event type representation matrix, each row of which corresponds to final event type representation $h_r^{T-1}$ at time step $T-1$. $\text{ConvFC}(\cdot)$ represents the 1D convolution layer and the fully connected layer.

TABLE 1
Statistics of datasets

| Dataset | #Entity | #Relation | Training set | Validation set | Test set | #Time step | Time granularity |
|---|---|---|---|---|---|---|---|
| ICEWS14-full | 7,128 | 230 | 74,845 | 8,514 | 7,371 | 365 | 24 hours |
| ICEWS14 | 205 | 171 | 35,665 | 7,369 | 7,598 | 365 | 24 hours |
| ICEWS18-full | 23,033 | 256 | 373,018 | 45,995 | 49,545 | 304 | 24 hours |
| ICEWS18 | 208 | 164 | 34,497 | 4,412 | 4,661 | 304 | 24 hours |
| GDELT18-full | 7,691 | 240 | 1,734,399 | 238,765 | 305,241 | 2,751 | 15 minutes |
| GDELT18 | 219 | 211 | 158,901 | 21,351 | 27,407 | 2,751 | 15 minutes |

The training objective of GTRL is to minimize the cross-entropy loss, which is formulated as:

$$\mathcal{L} = -\frac{1}{F}\sum_{i=1}^{F}\sum_{j=1}^{N_r} y_{i,j}\log P_{i,j} + (1-y_{i,j})\log(1-P_{i,j}) \quad (15)$$

where $F$ and $N_r$ are the numbers of samples in the training set and event types, respectively. $y_{i,j}$ denotes the label of event type $j$ for sample $i$, and $y_{i,j} \in \{0,1\}$. $P_{i,j}$ is the predicted probability of event type $j$ for sample $i$, calculated by Equation (14).

## 5 EXPERIMENTS

In this section, we justify the superiority of GTRL by conducting extensive experiments. Firstly, we introduce the experimental datasets and settings. Subsequently, we present the comparison with baselines, the ablation study, the parameter sensitivity analysis, and the case study.

### 5.1 Datasets

To evaluate the performance of GTRL, we conduct experiments on six real-world datasets, i.e., ICEWS14-full, ICEWS18-full, GDELT18-full, ICEWS14, ICEWS18, and GDELT18. The former three datasets are collected by Li et al. [32], covering periods from 2014/1/1 to 2014/12/31, from 2018/1/1 to 2018/10/31, and from 2018/1/1 to 2018/1/31, respectively. We preprocess them to obtain the last three datasets by setting the type field of entities as countries to conduct filtering, which only contain the records of country-specific events. Following [32], all datasets are split into the training set, validation set, and test set in chronological order. The detailed statistics of datasets are presented in Table 1.

### 5.2 Experimental Settings

GTRL is implemented in Python with PyTorch, and the source code is released on GitHub[1]. We evaluate GTRL on the event prediction task. The event prediction task in TKGs is to predict the probability of each event type between an entity pair at time step $T$, based on historical events at previous $T-1$ time steps.

For the number of entity groups in the entity group mapper, we select it from 10 to 20, and the optimal setting is 16. For the length of the historical window, we select it from 1 to 13, and the optimal setting is 7. For the number of layers in the hierarchical GCNs, we select it from 1 to 5, and the optimal setting is 2. The numbers of hidden units and layers in GRUs are set to 100 and 1, respectively. Following [7], the batch size is set to 16, and entity representations and event type representations are initialized with Xavier initialization [35]. The dimension of representations is set to 100. It is worth noting that, matrix **M** is randomly initialized and would be optimized during the training process. Adam [38] is adopted as the optimizer, and the initial learning rates for the parameters of mapping matrix **M** and other parameters are set to 0.05 and 0.001, respectively.

[1] https://github.com/xt-55/GTRL

Following [7], we choose the best model based on the early stopping strategy, which will stop training if the validation loss does not decrease for 3 consecutive epochs. To avoid the impact of random factors, we report the average results after evaluating each method five times.

Given an entity pair, we rank all event types regarding the probability obtained by Equation (14).

We use Hits@$k$ and MRR (Mean Reciprocal Rank) as our evaluation metrics. Hits@$k$ refers to the percentage of correct event types ranked in the top $k$. MRR refers to the average reciprocal rank of correct event types. Higher Hits@$k$ and MRR demonstrate better performance. The Hits@$k$ and MRR are formulated as:

$$\text{Hits@}k = \frac{1}{\sum_{j\in N}|\mathcal{L}_j|}\sum_{j\in N}\sum_{q\in\mathcal{L}_j} I(\text{rank}_q \leq k) \quad (16)$$

$$\text{MRR} = \frac{1}{\sum_{j\in N}|\mathcal{L}_j|}\sum_{j\in N}\sum_{q\in\mathcal{L}_j}\frac{1}{\text{rank}_q} \quad (17)$$

where $\mathcal{L}_j$ represents the true label set for sample $j$ and $|\mathcal{L}_j|$ is the number of true labels. $I(\cdot)$ is the indicator function. $\text{rank}_q$ is the rank of $q$ in the prediction.

### 5.3 Comparison with Baselines

To justify the superiority of GTRL, we compare it with static KG representation learning methods, i.e., TransE [1], DistMult [29], R-GCN [20], and CompGCN [24]; TKG representation learning methods, i.e., TTransE [14], TNT-ComplEx [13], TeMP [26], RE-NET [10], Glean [7], RE-GCN [32], and TITer [8]. The detailed descriptions of baselines are as follows:

TABLE 2
Performances (in percentage) of GTRL and the compared methods on GDELT18 and GDELT18-full (mean ± std). * indicates that GTRL is statistically superior to the compared method according to pairwise t-test at a 95% significance level. The best results are bolded and the second-best results are underlined

|  | GDELT18 | | | | GDELT18-full | | | |
| --- | --- | --- | --- | --- | --- | --- | --- | --- |
| Method | MRR | Hits@1 | Hits@3 | Hits@10 | MRR | Hits@1 | Hits@3 | Hits@10 |
| TransE (NIPS 2013) | 17.17 ± 0.21* | 2.98 ± 0.24* | 19.01 ± 0.61* | 31.18 ± 0.36* | 19.04 ± 0.11* | 7.83 ± 0.11* | 19.25 ± 0.02* | 33.56 ± 0.02* |
| DisMult (ICLR 2015) | 20.16 ± 0.01* | 13.46 ± 0.32* | 21.18 ± 0.01* | 33.27 ± 0.01* | 22.34 ± 0.01* | 15.77 ± 0.22* | 24.11 ± 0.03* | 35.25 ± 0.01* |
| R-GCN (ESWC 2018) | 23.74 ± 0.32* | 19.05 ± 0.01* | 26.99 ± 0.01* | 37.19 ± 0.29* | 25.16 ± 0.22* | 21.22 ± 0.01* | 28.94 ± 0.05* | 37.21 ± 0.49* |
| CompGCN (ICLR 2020) | 23.98 ± 0.01* | 19.88 ± 0.42* | 28.14 ± 0.01* | 37.96 ± 0.01* | 25.98 ± 0.05* | 21.89 ± 0.32* | 29.74 ± 0.01* | 38.96 ± 0.01* |
| TTransE (WWW 2018) | 9.84 ± 0.22* | 12.46 ± 0.01* | 17.59 ± 0.54* | 20.38 ± 0.01* | 11.04 ± 0.11* | 11.83 ± 0.11* | 19.25 ± 0.02* | 29.56 ± 0.02* |
| TNTComplEx (ICLR 2020) | 22.17 ± 0.73* | 18.15 ± 0.52* | 27.28 ± 0.01* | 35.37 ± 0.44* | 19.47 ± 0.01* | 15.69 ± 0.11* | 23.86 ± 0.01* | 40.43 ± 0.11* |
| RE-NET (EMNLP 2020) | 41.89 ± 0.01* | 37.74 ± 0.36* | 43.55 ± 0.21* | 49.69 ± 0.01* | 43.12 ± 0.11* | 36.72 ± 0.11* | 43.96 ± 0.11* | 48.59 ± 0.01* |
| Glean (KDD 2020) | 38.11 ± 0.32* | 30.29 ± 0.01* | 41.26 ± 0.01* | 43.35 ± 0.21* | 35.14 ± 0.11* | 26.29 ± 0.51* | 40.28 ± 0.33* | 45.37 ± 0.47* |
| TeMP (EMNLP 2020) | 43.23 ± 0.24* | 37.97 ± 0.01* | 44.05 ± 0.01* | 49.89 ± 0.01* | 39.23 ± 0.24* | 33.37 ± 0.01* | 42.45 ± 0.01* | 51.49 ± 0.01* |
| RE-GCN (SIGIR 2021) | 37.56 ± 0.01* | 35.09 ± 0.61* | 37.34 ± 0.11* | 42.44 ± 0.63* | 38.01 ± 0.01* | 29.53 ± 0.61* | 37.54 ± 0.11* | 48.44 ± 0.63* |
| TITer (EMNLP 2021) | 44.55 ± 0.28* | 38.08 ± 0.01* | 44.76 ± 0.01* | 49.98 ± 0.11* | 39.54 ± 0.28* | 31.52 ± 0.01* | 43.76 ± 0.01* | 47.47 ± 0.11* |
| **GTRL (ours)** | **49.33 ± 0.21** | **40.45 ± 0.01** | **50.39 ± 0.01** | **57.74 ± 0.01** | **40.13 ± 0.11** | **37.45 ± 0.07** | **47.39 ± 0.04** | **58.74 ± 0.05** |
| **Improvement** | **10.73%** | **6.22%** | **12.58%** | **15.53%** | **1.49%** | **1.99%** | **7.80%** | **14.08%** |

TABLE 3
Performances (in percentage) of GTRL and the compared methods on ICEWS18 and ICEWS18-full (mean ± std). * indicates that GTRL is statistically superior to the compared method according to pairwise t-test at a 95% significance level. The best results are bolded and the second-best results are underlined

|  | ICEWS18 | | | | ICEWS18-full | | | |
| --- | --- | --- | --- | --- | --- | --- | --- | --- |
| Method | MRR | Hits@1 | Hits@3 | Hits@10 | MRR | Hits@1 | Hits@3 | Hits@10 |
| TransE (NIPS 2013) | 19.76 ± 0.21* | 3.02 ± 0.01* | 23.74 ± 0.18* | 35.22 ± 0.08* | 20.04 ± 0.11* | 8.83 ± 0.11* | 23.25 ± 0.02* | 36.56 ± 0.02* |
| DisMult (ICLR 2015) | 22.27 ± 0.18* | 15.07 ± 0.32* | 29.97 ± 0.76* | 37.08 ± 0.09* | 24.21 ± 0.11* | 17.19 ± 0.32* | 31.04 ± 0.43* | 37.95 ± 0.09* |
| R-GCN (ESWC 2018) | 24.96 ± 0.02* | 19.77 ± 0.24* | 32.29 ± 0.23* | 40.73 ± 0.22* | 26.49 ± 0.02* | 21.15 ± 0.14* | 34.56 ± 0.27* | 42.53 ± 0.22* |
| CompGCN (ICLR 2020) | 25.47 ± 0.32* | 20.04 ± 0.01* | 34.18 ± 0.05* | 41.12 ± 0.27* | 27.17 ± 0.42* | 22.24 ± 0.05* | 35.16 ± 0.03* | 43.12 ± 0.21* |
| TTransE (WWW 2018) | 10.29 ± 0.03* | 11.04 ± 0.01* | 18.89 ± 0.08* | 23.74 ± 0.31* | 11.96 ± 0.24* | 13.97 ± 0.11* | 12.79 ± 0.01* | 24.33 ± 0.14* |
| TNTComplEx (ICLR 2020) | 22.98 ± 0.22* | 17.68 ± 0.25* | 30.98 ± 0.01* | 38.02 ± 0.23* | 21.85 ± 0.14* | 16.86 ± 0.39* | 25.64 ± 0.11* | 41.86 ± 0.25* |
| RE-NET (EMNLP 2020) | 42.05 ± 0.21* | 37.87 ± 0.37* | 44.78 ± 0.27* | 52.43 ± 0.01* | 42.25 ± 0.01* | 33.81 ± 0.24* | 44.98 ± 0.33* | 52.72 ± 0.61* |
| Glean (KDD 2020) | 39.58 ± 0.01* | 35.26 ± 0.58* | 43.44 ± 0.22* | 48.49 ± 0.04* | 37.11 ± 0.22* | 34.15 ± 0.04* | 42.56 ± 0.61* | 47.35 ± 0.61* |
| TeMP (EMNLP 2020) | 43.08 ± 0.23* | 38.07 ± 0.22* | 45.18 ± 0.01* | 53.03 ± 0.58* | 43.24 ± 0.24* | 38.77 ± 0.01* | 45.04 ± 0.01* | 55.94 ± 0.01* |
| RE-GCN (SIGIR 2021) | 44.27 ± 0.48* | 38.35 ± 0.01* | 44.75 ± 0.48* | 49.25 ± 0.11* | 41.56 ± 0.01* | 37.59 ± 0.61* | 44.34 ± 0.11* | 57.42 ± 0.63* |
| TITer (EMNLP 2021) | 45.07 ± 0.01* | 38.85 ± 0.23* | 46.44 ± 0.22* | 49.79 ± 0.45* | 45.44 ± 0.28* | 39.78 ± 0.01* | 48.77 ± 0.01* | 58.73 ± 0.11* |
| **GTRL (ours)** | **51.29 ± 0.21** | **43.14 ± 0.21** | **52.54 ± 0.23** | **59.73 ± 0.28** | **46.35 ± 0.11** | **40.95 ± 0.03** | **51.19 ± 0.21** | **60.18 ± 0.26** |
| **Improvement** | **13.80%** | **11.04%** | **13.14%** | **12.63%** | **2.00%** | **2.94%** | **4.96%** | **2.47%** |

TABLE 4
Performances (in percentage) of GTRL and the compared methods on ICEWS14 and ICEWS14-full (mean ± std). * indicates that GTRL is statistically superior to the compared method according to pairwise t-test at a 95% significance level. The best results are bolded and the second-best results are underlined

|  | ICEWS18 | | | | ICEWS18-full | | | |
| --- | --- | --- | --- | --- | --- | --- | --- | --- |
| Method | MRR | Hits@1 | Hits@3 | Hits@10 | MRR | Hits@1 | Hits@3 | Hits@10 |
| TransE (NIPS 2013) | 19.98 ± 0.11* | 4.11 ± 0.24* | 32.08 ± 0.08* | 39.88 ± 0.41* | 20.88 ± 0.11* | 9.23 ± 0.11* | 24.25 ± 0.02* | 39.56 ± 0.02* |
| DisMult (ICLR 2015) | 23.12 ± 0.34* | 16.02 ± 0.01* | 33.83 ± 0.07* | 45.01 ± 0.44* | 25.26 ± 0.11* | 18.23 ± 0.52* | 34.03 ± 0.23* | 39.95 ± 0.23* |
| R-GCN (ESWC 2018) | 25.08 ± 0.08* | 19.87 ± 0.37* | 35.14 ± 0.21* | 47.33 ± 0.07* | 27.74 ± 0.02* | 22.15 ± 0.14* | 36.88 ± 0.24* | 42.79 ± 0.22* |
| CompGCN (ICLR 2020) | 25.94 ± 0.41* | 20.03 ± 0.51* | 35.79 ± 0.33* | 47.73 ± 0.01* | 28.56 ± 0.42* | 25.24 ± 0.04* | 37.26 ± 0.03* | 43.92 ± 0.21* |
| TTransE (WWW 2018) | 11.79 ± 0.53* | 13.24 ± 0.01* | 19.97 ± 0.27* | 24.88 ± 0.56* | 23.79 ± 0.03* | 14.24 ± 0.51* | 29.17 ± 0.07* | 34.56 ± 0.05* |
| TNTComplEx (ICLR 2020) | 23.58 ± 0.01* | 18.06 ± 0.61* | 34.82 ± 0.17* | 45.45 ± 0.36* | 25.12 ± 0.73* | 18.15 ± 0.22* | 30.15 ± 0.01* | 45.37 ± 0.41* |
| RE-NET (EMNLP 2020) | 43.27 ± 0.48* | 38.97 ± 0.35* | 47.08 ± 0.01* | 55.19 ± 0.28* | 45.77 ± 0.28* | 37.98 ± 0.03* | 49.07 ± 0.07* | 58.87 ± 0.24* |
| Glean (KDD 2020) | 40.24 ± 0.01* | 35.62 ± 0.39* | 45.48 ± 0.53* | 50.09 ± 0.08* | 42.20 ± 0.71* | 36.86 ± 0.59* | 47.68 ± 0.43* | 52.39 ± 0.06* |
| TeMP (EMNLP 2020) | 44.17 ± 0.11* | 39.37 ± 0.18* | 47.78 ± 0.06* | 55.66 ± 0.08* | 46.04 ± 0.24* | 39.07 ± 0.01* | 49.84 ± 0.01* | 59.74 ± 0.01* |
| RE-GCN (SIGIR 2021) | 41.76 ± 0.61* | <u>39.67 ± 0.05</u>* | 45.37 ± 0.01* | 51.74 ± 0.11* | 45.56 ± 0.01* | 38.09 ± 0.61* | 50.37 ± 0.11* | <u>62.44 ± 0.63</u> |
| TITer (EMNLP 2021) | <u>44.86 ± 0.31</u>* | 39.37 ± 0.21* | <u>48.84 ± 0.06</u>* | <u>55.79 ± 0.51</u>* | <u>46.12 ± 0.28</u> | <u>39.08 ± 0.01</u>* | <u>50.76 ± 0.01</u>* | 60.39 ± 0.11* |
| **GTRL (ours)** | **51.95 ± 0.21** | **44.31 ± 0.33** | **54.09 ± 0.11** | **65.39 ± 0.32** | **46.25 ± 0.33** | **40.11 ± 0.53** | **51.09 ± 0.44** | **65.79 ± 0.27** |
| **Improvement** | **15.80%** | **11.70%** | **10.75%** | **17.21%** | **0.28%** | **2.64%** | **0.65%** | **5.37%** |

TABLE 5
Comparisons about the training and inference efficiency of different methods (The best results are bolded and the second-best results are underlined)

|  | ICEWS14 | | | | | ICEWS14-full | | | | |
| --- | --- | --- | --- | --- | --- | --- | --- | --- | --- | --- |
| Method | Total training time | Inference time | MRR | Hits@1 | Hits@10 | Total training time | Inference time | MRR | Hits@1 | Hits@10 |
| RE-NET | 12.08h | 167s | 43.27 | 38.97 | 55.19 | 15.69h | 203s | 45.77 | 37.98 | 58.87 |
| Glean | **3.12h** | <u>30s</u> | 40.24 | 35.62 | 50.09* | **4.55h** | <u>39s</u> | 42.20 | 36.86 | 52.39 |
| TeMP | 6.32h | 87s | 44.17 | 39.37 | 55.66 | 7.83h | 107s | 46.04 | 39.07 | 59.74 |
| RE-GCN | 4h | 18s | 41.76 | <u>39.67</u> | 51.74 | 5.21h | 36s | 45.56 | 38.09 | <u>62.44</u> |
| TITer | 13.89h | 160s | <u>44.86</u> | 39.37 | <u>55.79</u> | 16.28h | 182s | <u>46.12</u> | <u>39.08</u> | 60.39 |
| GTRL | <u>3.78h</u> | **12s** | **51.95** | **44.31** | **65.39** | <u>4.69h</u> | **20s** | **46.25** | **40.11** | **65.79** |

**Static KG representation learning methods:**

**TransE** [1] keeps the tail entity representation close to the sum of the head entity representation and the relation representation.

**DistMult** [29] models the semantic correlations between entities and relations by a bilinear model and employs diagonal matrices to represent relations.

**R-GCN** [20] obtains entity representations by learning different weights for relations and utilizing GCNs to model the influence of neighbor entities.

**CompGCN** [24] learns entity representations and relation representations by introducing entity-relation composition operations.

**TKG representation learning methods:**

**TTransE** [14] extends TransE [1], which models timestamps as corresponding representations.

**TNTComplEx** [13] extends ComplEx [31], which models timestamps as the order of tensor decomposition.

**TeMP** [26] uses R-GCN to model the influence of neighbor entities and employs a frequency-based gating GRU to model the temporal dependency among inactive events.

**RE-NET** [10] uses GCNs to model the influence of neighbor entities, employs RNNs to model the temporal dependency among events, and obtains the joint probability distribution of all events in an autoregressive way.

**Glean** [7] uses CompGCN to model the influence of neighbor entities and employs GRUs to model the temporal dependency among representations.

**RE-GCN** [32] is the SOTA method, which uses R-GCN to model the influence of neighbor entities and employs an auto-regressive GRU to model the temporal dependency among events.

**TITer** [8] is the SOTA method (based on paths), which incorporates temporal agent-based reinforcement learning to search paths and obtains entity representations by inductive mean.

For static KG representation learning methods, we conduct experiments based on triplets, ignoring timestamps. For the sake of equity, we compare all baselines based on the same experimental protocol with well-tuned hyper-parameters.

Tables 2, 3, and 4 show the average MRR, Hits@1, Hits@3, and Hits@10 of GTRL and baselines on GDELT18, GDELT18-full, ICEWS18, ICEWS18-full, ICEWS14, and ICEWS14-full, respectively, from which we can observe the following phenomena:

1) In static KG representation learning methods, CompGCN and R-GCN outperform TransE and DistMult, which demonstrates the effectiveness of modeling the graph structure information.

2) TKG representation learning methods RE-NET, Glean, TeMP, and RE-GCN outperform static KG representation learning methods R-GCN and CompGCN, which indicates that RE-NET, Glean, TeMP, and RE-GCN can obtain more accurate prediction results by modeling the temporal information.

3) In TKG representation learning methods, RE-NET, Glean, TeMP, and RE-GCN outperform TTransE and TNTComplEx, which demonstrates the effectiveness of modeling the temporal dependency.

4) As the SOTA method based on paths, TITer performs better than RE-NET, Glean, TeMP, and RE-GCN, which indicates that TITer can obtain more accurate prediction results by introducing paths to model the influence of distant entities.

5) GTRL surpasses RE-NET, Glean, TeMP, RE-GCN, and TITer, surpassing the best baseline by an average of 7.35%, 6.09%, 8.31%, and 11.21% in MRR, Hits@1, Hits@3, and Hits@10 on six datasets, respectively. The results justify the advantage of introducing the entity group modeling to capture the correlation between entities that are far apart and even unreachable. Note that the improvements of GTRL on ICEWS14, ICEWS18, and GDELT18 are superior to those on ICEWS14-full, ICEWS18-full, and GDELT18-full. This underscores GTRL's ability to model latent correlations effectively through the entity grouping mechanism, especially in sparse scenarios, which leads to a significant performance boost.

To evaluate the training and inference efficiency, we compare the training time, inference time, and performances of RE-NET, Glean, TeMP, RE-GCN, TITer, and GTRL on ICEWS14 and ICEWS14-full in Table 5. Glean trains the fastest in these methods, but gets relatively bad results. Compared with RE-NET, TeMP, RE-GCN, and TITer, GTRL runs the fastest and gets the best results.

Overall, comprehensively considering the consistent performance improvement and the training and inference efficiency, GTRL demonstrates the superiority over existing methods.

### 5.4 Ablation Study

To justify the advantage of introducing the entity group mapper to generate entity groups, we compare GTRL with three variants. The detailed descriptions of variants are as follows:

**GTRL-Single Graph (GTRL-S)**: We remove the entity group graph and only utilize the 2-layer CompGCN to model the entity graph.

**GTRL-K-Means (GTRL-K)**: We replace the mapping matrix of the entity group mapper with the grouping results of the K-means algorithm ($k = 16$) based on entity representations obtained by baseline CompGCN [24].

**GTRL-DNN (GTRL-D)**: We replace the entity group mapper with a specialized DNN, which utilizes entity representations as the input of the MLP followed by the softmax function to obtain the mapping matrix of the entity group mapper.

Fig. 3 shows the average Hits@1, Hits@10, and MRR of GTRL and its variants that employ different ways to generate entity groups on ICEWS14, from which we can observe the following phenomena:

1) GTRL-K outperforms GTRL-S, which indicates the effectiveness of constructing the entity group graph to model the influence of distant entities.

2) GTRL performs better than GTRL-K and GTRL-D. The results justify the advantage of introducing the entity groups from entities in a learning way. Though using DNNs to compute the mapping matrix may be beneficial in some cases, it may also introduce additional complexity and constraints that can affect the performance of the model. Directly parameterizing the mapping matrix can provide more flexibility to the model, resulting in better performance.

To justify the advantage of introducing the implicit correlation encoder to capture implicit correlations between pairwise entity groups, we compare GTRL with five variants. The detailed descriptions of variants are as follows:

**GTRL-Entity Graph (GTRL-EG)**: We construct the edge of the entity group graph based on whether there is an edge between corresponding entities on the entity graph, and set the intensity of the implicit correlation to 1.

**GTRL-Entity Graph with Intensity (GTRL-EG-I)**: Different from setting the intensity of the implicit correlation to 1 in GTRL-EG, we retain implicit correlation intensities.

**GTRL-Sparse Graph (GTRL-SG)**: We construct the entity group graph as the sparse graph based on the top $n$ strategy [3], and set top $n$ intensities to 1 for each entity group ($n = 3$), with the other intensities set to 0.

**GTRL-Sparse Graph with Intensity (GTRL-SG-I)**: Different from setting top $n$ intensities to 1 for each entity group in GTRL-SG, we retain top $n$ intensities for each entity group ($n = 3$).

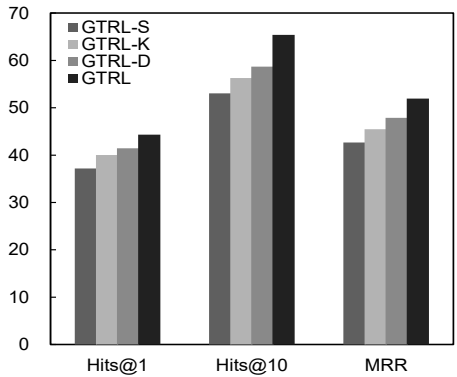
Fig. 3. Comparison of GTRL and its variants that employ different ways to generate entity groups.

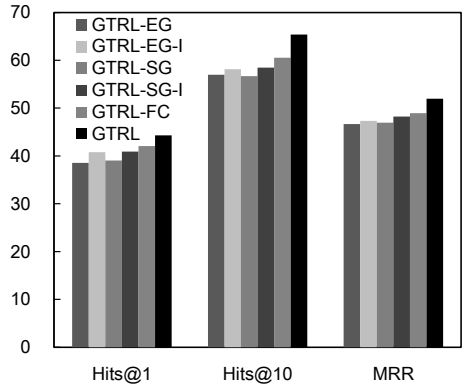
Fig. 4. Comparison of GTRL and its variants that employ different ways to capture implicit correlations between pairwise entity groups.

TABLE 6
The comparison (in percentage) of GTRL and its variants with different ways to perform the message passing (mean ± std). * indicates that GTRL is statistically superior to its variants according to pairwise t-test at a 95% significance level. The best results are in bold

| | ICEWS14 | | | ICEWS14-full | | |
|---|---|---|---|---|---|---|
| Method | MRR | Hits@1 | Hits@10 | MRR | Hits@1 | Hits@10 |
| GTRL-Chebyshev | 44.77 ± 0.22* | 40.04 ± 0.33* | 52.73 ± 0.31* | 42.15 ± 0.27* | 37.17 ± 0.33* | 55.79 ± 0.21* |
| GTRL-attention | 45.56 ± 0.01* | 41.09 ± 0.61* | 57.44 ± 0.03* | 44.76 ± 0.21* | 39.67 ± 0.18* | 56.74 ± 0.11* |
| GTRL | **51.95 ± 0.21** | **44.31 ± 0.33** | **65.39 ± 0.32** | **46.25 ± 0.33** | **40.11 ± 0.53** | **65.79 ± 0.27** |

TABLE 7
The comparison (in percentage) of GTRL and its variants with different ways to perform the message passing (mean ± std). * indicates that GTRL is statistically superior to its variants according to pairwise t-test at a 95% significance level. The best results are in bold

| | ICEWS14 | | | ICEWS14-full | | |
|---|---|---|---|---|---|---|
| Method | MRR | Hits@1 | Hits@10 | MRR | Hits@1 | Hits@10 |
| GTRL-vanilla RNNs | 46.59 ± 0.22* | 40.68 ± 0.31* | 57.28 ± 0.31* | 42.97 ± 0.25* | 37.56 ± 0.28* | 56.85 ± 0.27* |
| GTRL-LSTM | 47.86 ± 0.01* | 41.55 ± 0.28* | 59.76 ± 0.05* | 43.77 ± 0.21* | 37.75 ± 0.18* | 57.28 ± 0.11* |
| GTRL-self attention | 49.36 ± 0.01* | 42.32 ± 0.61* | 63.17 ± 0.03* | 45.04 ± 011* | 39.48 ± 0.11* | 59.65 ± 0.17* |
| GTRL | **51.95 ± 0.21** | **44.31 ± 0.33** | **65.39 ± 0.32** | **46.25 ± 0.33** | **40.11 ± 0.53** | **65.79 ± 0.27** |

TABLE 8
The comparison (in percentage) of GTRL and its variant with different frameworks to capture the temporal information and graph structure information (mean ± std). * indicates that GTRL is statistically superior to its variants according to pairwise t-test at a 95% significance level. The best results are in bold

| | ICEWS14 | | | ICEWS14-full | | |
|---|---|---|---|---|---|---|
| Method | MRR | Hits@1 | Hits@10 | MRR | Hits@1 | Hits@10 |
| GTRL-time | 50.31 ± 0.07* | 43.24 ± 0.21* | 64.11 ± 0.07* | 45.28 ± 014* | 39.69 ± 0.14* | 63.27 ± 0.14* |
| GTRL | **51.95 ± 0.21** | **44.31 ± 0.33** | **65.39 ± 0.32** | **46.25 ± 0.33** | **40.11 ± 0.53** | **65.79 ± 0.27** |

**GTRL-Fully Connected (GTRL-FC)**: We construct the entity group graph as the fully connected graph and set the intensity of the implicit correlation to 1.

Fig. 4 shows the average Hits@1, Hits@10, and MRR of GTRL and its variants that employ different ways to capture implicit correlations between pairwise entity groups on ICEWS14, from which the following phenomena can be observed:

1) GTRL-FC performs better than GTRL-EG, GTRL-EG-I, GTRL-SG, and GTRL-SG-I. The results indicate that by constructing the entity group graph as the fully connected graph, GTRL-FC can fully utilize the information between any pairwise entity groups, while GTRL-EG, GTRL-EG-I, GTRL-SG, and GTRL-SG-I construct the entity group graph as the non-fully connected graph, leading to information loss.

2) GTRL outperforms GTRL-FC, and GTRL-EG-I and GTRL-SG-I perform better than GTRL-EG and GTRL-SG, respectively. The results justify the advantage of introducing the intensity of the implicit correlation. By incorporating the intensity of the implicit correlation into the message aggregation, GTRL, GTRL-EG-I, and GTRL-SG-I can distinguish different implicit correlations.

To justify the advantage of introducing hierarchical GCNs to implement the information interaction on the entity group graph and the entity graph, we compare GTRL with two variants. The detailed descriptions of variants are as follows:

**GTRL-Chebyshev**: We replace the four-stage message passing mechanism with the Chebyshev spectral graph convolutional operator [46] from the entity graph to the entity group graph.

**GTRL-attention**: We replace the four-stage message passing mechanism with the one-way GAT [47] from the entity graph to the entity group graph.

Table 6 shows the average Hits@1, Hits@10, and MRR of GTRL and its variants that employ different ways to implement the information interaction on the entity group graph and the entity graph on ICEWS14 and ICEWS14-full, from which we can observe the following phenomena:

1) GTRL-attention outperforms GTRL-Chebyshev, which indicates the effectiveness of aggregating messages in an attentive way.

2) GTRL performs better than GTRL-attention. The results justify the advantage of introducing the four-stage message passing. Equipped with entity group representation updating, entity group graph convolution, entity representation updating, and entity graph convolution, GTRL can accomplish two-way message passing, which ensures comprehensive information interaction on the entity graph and the entity group graph.

To justify the advantage of introducing decay-aware GRUs to model the temporal dependency among representations, we compare GTRL with three variants. The detailed descriptions of variants are as follows:

**GTRL-vanilla RNNs**: We utilize vanilla RNNs to encode the temporal dependency among updated representations.

**GTRL-self attention**: We utilize self attention [19] to encode the temporal dependency among updated representations.

**GTRL-LSTM**: We utilize LSTM to encode the temporal dependency among updated representations.

Table 7 shows the average Hits@1, Hits@10, and MRR of GTRL and its variants that employ different ways to encode the temporal dependency on ICEWS14-full and ICEWS14, from which we can observe the following phenomena:

1) GTRL-self attention outperforms GTRL-vanilla RNNs and GTRL-LSTM, which indicates the importance of differentiating the effects of different representations.

2) GTRL performs better than GTRL-self attention. The results verify the effectiveness of the decay-aware GRUs. These decay-aware GRUs are designed to emphasize more recent events and gradually reduce the influence of older ones. This inherent time-sensitive nature allows for more nuanced and realistic modeling of how information becomes less relevant over time, leading to better results in scenarios where the relevance of events changes over time, while GTRL-self attention might give high weights to some events irrespective of their timestamps.

As discussed in [48], there are two frameworks, i.e., "time and GNN" and "time then GNN", to jointly capture the temporal information and graph structure information. To justify the advantage of introducing the "time and GNN" framework, we compare GTRL with GTRL-time. GTRL-time exploits the "time then GNN" framework by first applying RNN to capture the temporal information among representations, and then employing hierarchical GNN to capture graph structure information.

Table 8 shows the average Hits@1, Hits@10, and MRR of GTRL and its variant that exploits different frameworks to jointly capture the temporal information and graph structure information on ICEWS14 and ICEWS14-full. Experimental results show that, in the context of TKGs, the "time and GNN" framework, as implemented in GTRL, outperforms the "time then GNN" framework, as implemented in GTRL-time. The richness and variability of entity and edge types in TKGs make it challenging to extract all temporal information first before applying hierarchical GNNs. Armed with the entity group mapper and implicit correlation encoder, GTRL can capture latent relationships among entities in TKGs via the "time and GNN" framework.

## 5.5 Parameter Sensitivity Analysis

To study the impact of several important parameters, including the number of entity groups in the entity group mapper, the length of the historical window, and the number of layers in the hierarchical GCNs, we conduct the parameter sensitivity analysis on ICEWS14.

We evaluate the impact of the number of entity groups in the entity group mapper, varying it from 10 to 20 with the step of 2. Fig. 5 shows the evaluation results on ICEWS14, from which we can observe that with the increase of the number of entity groups, firstly the performance rises and then drops slightly. The best performance is achieved when there are 16 entity groups. This may be because the large entity group number brings challenges to the entity grouping. In addition, the number of edges on the entity group graph increases exponentially with the number of entity groups. When there are too many entity groups, dense edges may contain excessive noise that would reduce the capacity of representing entities.

We evaluate the impact of the length of the historical window, varying it from 1 to 13 with the step of 2. Fig. 6 shows the evaluation results on ICEWS14, from which we can observe that with the length of the historical window becoming longer, firstly the performance goes up and then goes down gradually. The best performance is

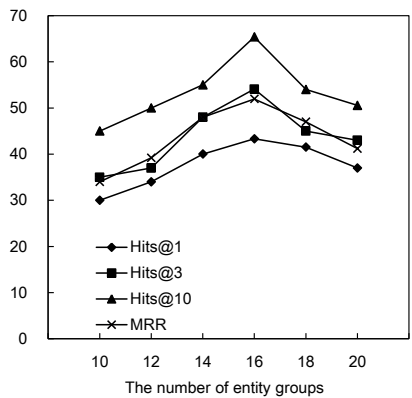

Fig. 5. Impact of the number of entity groups in the entity group mapper.

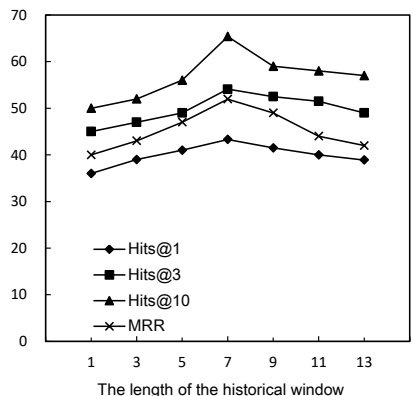

Fig. 6. Impact of the length of the historical window.

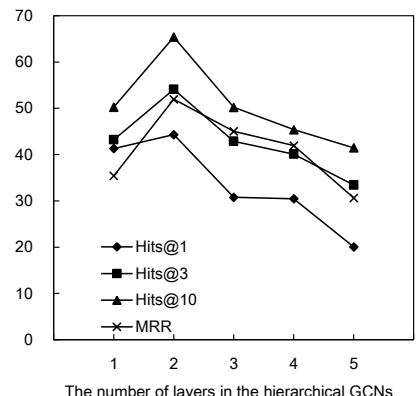

Fig. 7. Impact of the number of layers in the hierarchical GCNs.

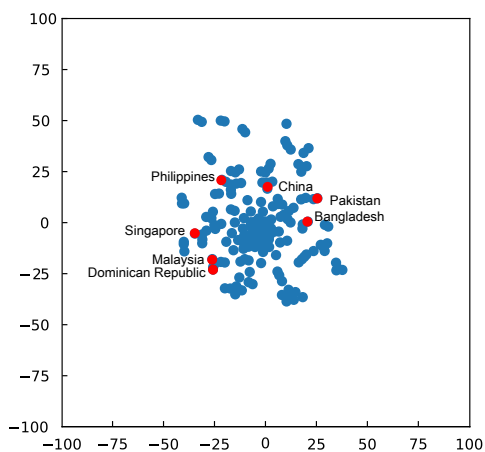

(a) RE-GCN

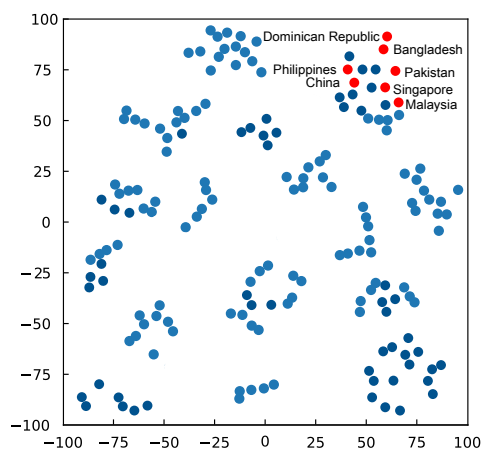

(b) GTRL

Fig. 8. Visualization of entity representations obtained in the last training historical window using t-SNE.

TABLE 9
Top 5 event types predicted by TITer and GTRL for several test samples in the test set of ICEWS18. Accurate prediction results are bolded

| Test sample | TITer Top 5 predicted event types | GTRL Top 5 predicted event types |
|---|---|---|
| (Japan, ?, China, 2018/10/25) | Make statement<br>**Make a visit**<br>Demand<br>Express intent to meet or negotiate<br>Host a visit | **Make a visit**<br>**Sign formal agreement**<br>**Engage in diplomatic cooperation**<br>Engage in negotiation<br>Praise or endorse |
| (China, ?, Japan, 2018/10/25) | Consult<br>Praise or endorse<br>Make optimistic comment<br>**Host a visit**<br>Appeal for diplomatic cooperation | **Sign formal agreement**<br>**Host a visit**<br>**Engage in diplomatic cooperation**<br>Make optimistic comment<br>**Express intent to engage in diplomatic cooperation** |
| (Japan, ?, China, 2018/10/26) | Praise or endorse<br>Make pessimistic comment<br>**Sign formal agreement**<br>Engage in diplomatic cooperation<br>Criticize or denounce | **Sign formal agreement**<br>**Express intent to engage in diplomatic cooperation**<br>Consult<br>**Make a visit**<br>**Appeal for easing of administrative sanctions** |
| (China, ?, Japan, 2018/10/26) | Make an appeal or request<br>**Sign formal agreement**<br>Express intent to meet or negotiate<br>Cooperate economically<br>Make optimistic comment | **Sign formal agreement**<br>**Praise or endorse**<br>**Express intent to cooperate**<br>Consult<br>**Engage in diplomatic cooperation** |

achieved when the length of the historical window is 7. The reason may be that when the length of the historical window is too long, the number of graphs that need to be processed will be large, which may cause the vanishing gradient problem in GRUs.

We evaluate the impact of the number of layers in the hierarchical GCNs, varying it from 1 to 5 with the step of 1. Fig. 7 shows the evaluation results on ICEWS14, from which we can observe that with the increase of the number of layers, firstly the performance increases and then degrades rapidly. The best performance is achieved when the number of layers is 2. The reason may be that the model complexity grows with the number of layers in hierarchical GCNs. When the number of layers is too large, the high model complexity will increase the risk of over-fitting.

### 5.6 Case Study

To intuitively reveal the superiority of GTRL, we perform several case studies.

In Fig. 8, we use the t-SNE method [36] to visualize entity representations in the last training historical window, obtained by RE-GCN [32] and GTRL on ICEWS18. We can find that entity representations learned by RE-GCN are distributed concentratedly. On the contrary, entity representations learned by GTRL are distributed widely. In addition, GTRL makes the representations of highly correlated entities distributed closer together. For example, the representations of countries that participate in the Belt and Road Initiative, e.g., "Pakistan", "Bangladesh", "Dominican Republic", "Philippines", "Malaysia", and "Singapore", are distributed close together. It indicates that by introducing the entity group modeling to capture the correlation between entities, GTRL can learn better entity representations.

Table 9 shows the top 5 predicted event types for several test samples in the test set of ICEWS18 obtained by TITer [8] and GTRL. On October 25, 2018, Japanese Prime Minister Shinzo Abe visited China again after seven years. Subsequently, on October 26, 2018, the first China-Japan Third-Party Market Cooperation Forum was held in Beijing, and 52 cooperation agreements were signed. Thus, the following four test samples, i.e., (Japan, ?, China, 2018/10/25), (China, ?, Japan, 2018/10/25), (Japan, ?, China, 2018/10/26), and (China, ?, Japan, 2018/10/26), are investigated. We can find that GTRL predicts more accurate event types than TITer and the accurately predicted event types are ranked higher than those of TITer. It demonstrates that by introducing the entity group modeling to capture the correlation between entities, GTRL can obtain more accurate prediction results.

Fig. 9 shows the proportions and Hits@1 of test samples with different shortest path lengths. We can observe that the performance of GTRL is stable in test samples with different shortest path lengths. GTRL slightly surpasses TITer in test samples whose shortest path lengths are between 1 and 3, and the performance gap between them is large in the other test samples. It justifies that by introducing the entity group mapper, implicit correlation encoder, and hierarchical GCNs, GTRL can capture the correlation between entities that are far apart and even unreachable by stacking only 2 layers, while TITer only models 1-hop, 2-hop, and 3-hop paths, which cannot capture the rich information from longer paths.

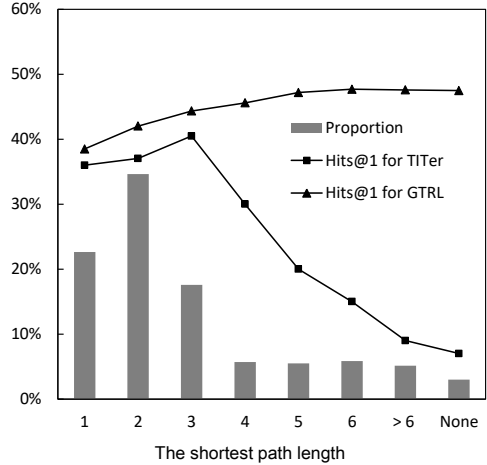

Fig. 9. The proportions and Hits@1 of test samples with different shortest path lengths. ("None" denotes no reachable paths for an entity pair.)

## 6 CONCLUSIONS AND FUTURE WORK

Capturing the correlation between entities and exploiting the information of both adjacent and distant entities are crucial for TKG representation learning. To this end, we propose GTRL, an entity group-aware temporal knowledge graph representation learning method, which incorporates the entity group modeling to capture the correlation between entities that are far apart and even unreachable. Specifically, the entity group mapper is proposed to generate entity groups from entities. The implicit correlation encoder is introduced to capture implicit correlations between any pairwise entity groups. In addition, the hierarchical GCNs are exploited to model the influence of both adjacent and distant entities. We conduct comprehensive experiments on six real-world datasets and experimental results demonstrate the superiority of GTRL.

In the future, we will extend this work in the following directions. First, we will investigate how to leverage more information, e.g., text descriptions and geographical locations, to discover multi-source correlations between entities. Second, we also plan to introduce a multi-level group modeling method to capture more complicated correlations between entities. Third, we will investigate advanced techniques, e.g., low-rank approximations, to achieve a tradeoff between performance and model complexity.

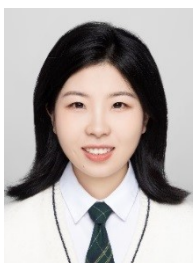
Xing Tang received her Ph.D. degree in computer science from Zhejiang University, China, in 2023. Her research interests include knowledge graph, data mining, and deep learning.

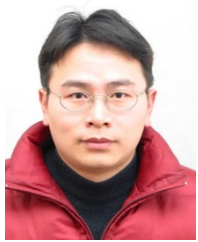
Ling Chen received his B.S. and Ph.D. degrees in computer science from Zhejiang University, China, in 1999 and 2004, respectively. He is currently a professor with the College of Computer Science and Technology, Zhejiang University, China. His research interests include ubiquitous computing and data mining.